\documentclass[lettersize,journal]{IEEEtran}
\usepackage{amsmath,amsfonts}
\usepackage{algorithmic}
\usepackage{algorithm}
\usepackage{array}
\usepackage[caption=false,font=normalsize,labelfont=sf,textfont=sf]{subfig}
\usepackage{textcomp}
\usepackage{stfloats}
\usepackage{url}
\usepackage{verbatim}
\usepackage{graphicx}
\usepackage{cite}
\usepackage{xcolor}
\usepackage{multirow} 
\usepackage{booktabs}
\usepackage{hyperref}

\begin{document}

\title{Speculative Decoding Reimagined for Multimodal Large Language Models}

\author{
Luxi Lin,
Zhihang Lin,
Zhanpeng Zeng,
and Rongrong Ji, Senior Member, IEEE
\thanks{Luxi Lin and Zhihang Lin contributed equally to this work.}
\thanks{L. Lin, Z. Lin, Z. Zeng, and R. Ji are with the Key Laboratory of Multimedia Trusted Perception and Efficient Computing, Ministry of Education of China, School of Informatics, Xiamen University, Xiamen 361005, China.}
\thanks{Z. Lin is also with the Shanghai Innovation Institute, China.}
\thanks{Corresponding author: Rongrong Ji.}
}



\maketitle

\begin{abstract}
This paper introduces Multimodal Speculative Decoding (MSD) to accelerate Multimodal Large Language Models (MLLMs) inference.
Speculative decoding (SD) accelerates Large Language Models (LLMs) without sacrificing accuracy, but existing SD methods for MLLMs fail to achieve comparable speedups. 
To address this, we reimagine speculative decoding specifically for MLLMs.
Our analysis of MLLM characteristics reveals three key design principles for MSD:
(1) Text and visual tokens have fundamentally different characteristics and need to be processed separately during drafting.
(2) Visual inputs to the draft model are highly redundant and can be compressed into a more informative token set.
(3) Both language modeling and visual perception capabilities are crucial for the draft model.
MSD instantiates these principles by decoupling text and visual tokens in the draft model, allowing each to be handled based on its own characteristics.
It further equips the draft model with a lightweight visual token reduction module that constructs a compact, informative visual representation before drafting.
MSD also adopts a two-stage training strategy: it first learns language modeling from text-only instructions and then builds visual perception from multimodal data.
Experiments show that MSD boosts inference speed by up to $2.91\times$ for LLaVA-1.6-7B and up to $3.24\times$ for LLaVA-1.6-13B on multimodal benchmarks.
The source code is publicly available at 
\url{https://github.com/Lyn-Lucy/MSD}.

\end{abstract}

\begin{IEEEkeywords}
Multimodal large language models, speculative decoding, inference acceleration
\end{IEEEkeywords}

\section{Introduction}
\label{sec:intro}
\IEEEPARstart{M}{ultimodal} large language models (MLLMs)~\cite{liu2023llava1.5,li2023blip2,liu2024llavanext,bai2025qwen25vl} have made significant strides in cross-modal applications~\cite{goyal2017vqav2,chen2015coco,achiam2023gpt,team2023gemini}.
MLLMs usually consist of a vision encoder, a projection layer, and a large language model (LLM).
Through visual instruction tuning, MLLMs achieve remarkable performance in multimodal tasks~\cite{liu2023llava1.5}.
However, MLLMs inherit the large parameter size of LLMs, leading to substantial computational overhead and slow inference speed.
These limitations have become critical bottlenecks for real-world deployment applications where low latency and high throughput are essential requirements for practical usage.

To optimize the inference efficiency of MLLMs, researchers have proposed various methods, such as pruning~\cite{chen2023diffrate,wu2023ppt}, quantization~\cite{wang2024quant,li2025mbq}, distillation~\cite{wang2024pruning}, and efficient architecture design~\cite{lin2024moellava}.
However, these methods often compromise accuracy, which is undesirable for tasks requiring consistent and high-quality outputs.
Therefore, accelerating MLLM inference without sacrificing accuracy remains a key challenge.

Speculative decoding (SD) has been proven to be an effective lossless inference acceleration method in LLMs~\cite{leviathan2023spd}.
SD uses a lightweight draft model to generate multiple candidate tokens, which are then verified in parallel by a target model (LLM).
SD breaks the paradigm of generating one token at a time and can significantly speed up the inference of LLMs.
As the main computational overhead of MLLMs comes from the LLM portion, SD can also be extended to MLLMs.
Recently, Gagrani \emph{et al.}~\cite{gagrani2024spd_mllm} conduct an initial exploration of SD in MLLMs, showing that draft models trained on text-only datasets can accelerate MLLM inference.
However, the average acceptance length of their draft model is only 2.5, which is far from the average acceptance length (5) of the state-of-the-art draft model in LLMs~\cite{li2024eagle2}.
This gap indicates that the existing SD methods for MLLMs have not fully utilized the potential of SD.
Therefore, designing a more effective and efficient SD method to significantly accelerate MLLM inference remains an important open and pressing research challenge for the community.

\begin{table}[tbp]
    \centering
    {
    \caption{\label{tab:Ablation_level}
    Average acceptance length of draft models with different predictive objectives on three multimodal tasks: ChartQA~\cite{masry2022chartqa}, TextVQA~\cite{Singh2019textvqa}, and HallusionBench~\cite{guan2024hallusionbench}, using LLaVA-1.5-7B~\cite{liu2023llava1.5} as the target model. Both draft models use the same architecture and number of parameters. ``Hallusion'' denotes HallusionBench in all tables.
    }
    }
    
    \resizebox{0.9\columnwidth}{!}{%
    \begin{tabular}{@{}lcccc@{}}
    \toprule
    \textbf{Type} & \textbf{ChartQA} & \textbf{TextVQA} & \textbf{Hallusion} & \textbf{Average} \\ 
    \midrule
     Token-level & 2.01 & 2.42 & 2.12 & 2.18 \\
     Feature-level & 3.45 & 3.81 & 3.72 & 3.66 \\
    \bottomrule
    \end{tabular}
    }
\end{table}

Based on the predictive objective of draft models, speculative decoding methods can be divided into two categories: token-level and feature-level. 
Token-level methods like Medusa~\cite{cai2024medusa} and Lookahead~\cite{fu2024lade} directly predict the next token, while feature-level methods like Eagle~\cite{li2024eagle} predict the feature of the next token. 
Feature-level methods achieve a higher acceleration ratio and average acceptance length than token-level methods in LLMs.
To validate whether this conclusion still holds in MLLMs, we conduct an ablation study comparing the average acceptance length of token-level and feature-level SD methods in MLLMs. 
As shown in Table~\ref{tab:Ablation_level}, the feature-level SD consistently outperforms the token-level SD in MLLMs.
However, the average acceptance length of feature-level SD in MLLMs still lags behind that of LLMs, indicating that existing SD methods cannot be directly applied to MLLMs. 
Thus, we need to reimagine the design of SD for MLLMs to fully leverage its acceleration potential.

In this paper, we introduce Multimodal Speculative Decoding (MSD), a novel speculative decoding method tailored for MLLMs.
We first conduct an in-depth analysis of the characteristics of MLLMs in the methods section, leading to three key design principles for MSD:
1) The draft model should decouple the processing of text and visual tokens, allowing each to be processed in a manner that aligns with its unique modality characteristics.
2) Visual inputs to the draft model are highly redundant, so the draft model should operate on a compact and informative visual representation rather than all raw visual tokens.
3) The draft model should be trained on both text-only and vision instruction datasets, ensuring strong language modeling and visual perception capabilities.
For the first principle, MSD concatenates the features of text tokens with next-token embeddings, while directly feeding visual token embeddings into the draft model. This modality-specific input construction preserves the distinct information carried by textual and visual representations and provides more appropriate inputs for multimodal drafting. For the second principle, MSD applies a lightweight visual token reduction module to compress the visual inputs seen by the draft model. By clustering redundant visual embeddings into compact representatives, it reduces draft computation while preserving candidate quality, thereby improving end-to-end speculative decoding efficiency. For the third principle, MSD employs a progressive two-stage training strategy for the draft model. It first strengthens language modeling on text-only data and then gradually introduces multimodal data, allowing the capacity-limited single-layer draft model to acquire visual representations while preserving language ability.

We evaluate MSD on various multimodal tasks, including visual question answering~\cite{goyal2017vqav2}, document understanding~\cite{masry2022chartqa}, and comprehensive benchmarking tasks~\cite{Liu2023mmbench}.
Experiments show that MSD boosts inference speed by up to $2.91\times$ for LLaVA-1.6-7B and up to $3.24\times$ for LLaVA-1.6-13B on multimodal benchmarks, further demonstrating its strong effectiveness and practical efficiency.

In summary, our main contributions are as follows:
\begin{itemize}
    \item We conduct an in-depth analysis of MLLM characteristics, from which we derive three key design principles for building effective draft models for MLLMs. These design principles provide valuable insights for future research on speculative decoding in MLLMs.
    
    \item Based on these principles, we propose MSD, a novel speculative decoding method for MLLMs that decouples the processing of text and visual tokens, incorporates a lightweight visual token reduction module for the draft model, and employs a two-stage training strategy.
    
    \item Extensive experiments demonstrate that MSD significantly accelerates MLLM inference and achieves state-of-the-art performance on multimodal benchmarks.
\end{itemize}

\section{Related work}
\label{sec:related}
\if false
\subsection{Efficient inference for MLLMs}
Recent research on enhancing the inference efficiency of Multimodal Large Language Models (MLLMs) primarily explores two paradigms: lossy acceleration and lossless acceleration. 
Lossy acceleration methods improve efficiency by sacrificing certain model capabilities. For example, architectural simplification approaches, such as TinyLLaVA and LLaVA-Phi, replace large backbones with compact alternatives. Token compression and pruning techniques, like TokenPacker for feature condensation and FastV/VTW for discarding low-relevance visual tokens in deeper layers, enhance efficiency. While effective, these approaches risk information loss and performance degradation. 
In contrast, lossless acceleration optimizes computation without compromising accuracy. It includes efficient attention methods like FlashAttention and PageAttention to reduce memory and computation costs, as well as speculative decoding, where a draft model proposes tokens and a target model verifies them. 
This article primarily focuses on speculative sampling within the realm of lossless acceleration.

\subsection{Speculative decoding}
Speculative decoding~\cite{leviathan2023spd} has proven to be an effective technique for accelerating inference in LLMs. This method utilizes a draft model to generate multiple candidate tokens, which are then concurrently validated by a target model, significantly speeding up the decoding stage.
Various studies focus on designing specialized draft models to enhance decoding efficiency further.
Medusa\cite{cai2024medusa} introduces multiple prediction heads that can predict several subsequent tokens simultaneously.
Eagle\cite{li2024eagle,li2024eagle2} and HASS\cite{zhang2024hass} employ a single-layer Transformer to predict the hidden state of the next token.
Additional research has explored the use of the target model or its sub-networks as the draft model.
SPACE\cite{yi2024space} achieves parallel token generation and validation through semi-autoregressive fine-tuning and automatic correction decoding algorithms.
Self-Speculative Decoding\cite{zhang2024draft} selectively skips intermediate layers, utilizing its own model to generate draft tokens.
While these methods have made significant progress in the LLMs field, speculative decoding has not been fully explored in MLLMs. In LLMs, Eagle is widely regarded as one of the most advanced techniques. This paper builds on the Eagle method, proposing targeted improvements to better address the challenges of multimodal scenarios.
\fi

\subsection{Speculative decoding}
Speculative decoding~\cite{leviathan2023spd} utilizes a draft model to generate multiple candidate tokens, which are then validated by a target model, significantly speeding up LLM decoding.
For example, Medusa~\cite{cai2024medusa} uses multiple prediction heads to generate several tokens at once. Eagle~\cite{li2024eagle,li2024eagle2} and HASS~\cite{zhang2024hass} employ a single-layer Transformer to predict the hidden state of the next token.
However, speculative decoding remains largely unexplored in MLLMs.
Gagrani \emph{et al.}~\cite{gagrani2024spd_mllm} argued that text-only trained draft models outperform multimodal ones. ViSpec~\cite{vispec} introduces vision adaptors and a global visual feature, yet its heavy compression of all visual tokens into a single representation restricts fine-grained visual use and adds extra computational overhead.
Recent studies further explore dynamic tree-based drafting~\cite{huo2025specllava}, lightweight draft branches~\cite{shao2025twig}, systematic benchmarking of multimodal speculative decoding~\cite{shen2026mmspec}, and latent-aware semi-autoregressive drafting~\cite{wang2025flash}.
These concurrent or subsequent works were unavailable when our experiments were conducted and are therefore discussed qualitatively rather than included as quantitative baselines.
These studies mark important early progress toward enabling speculative decoding in multimodal models; nevertheless, robust cross-modal interaction modeling and effective capability learning in draft models still remain significant and non-trivial challenges.

\section{Background}
\label{sec:prelim}

\subsection{Multimodal Large Language Model}

Representative MLLMs include LLaVA~\cite{liu2023llava1.5}, Qwen-VL~\cite{Qwen-VL}, and InternVL~\cite{chen2024internvl}. We use LLaVA as an example for input illustration.

\begin{figure}[t]
    \centering
    \includegraphics[width=0.45\textwidth]{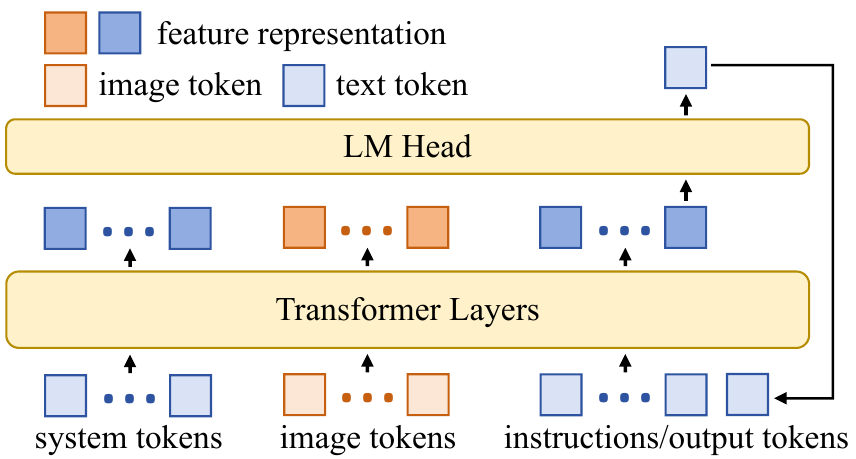}
    \caption{
        Input illustration of MLLMs. The input to MLLMs typically comprises two distinct types of tokens: textual tokens and visual tokens
        }
    \label{fig:mllm_input}
\end{figure}

As shown in Figure~\ref{fig:mllm_input}, the input consists of text tokens $
\{t_1,\dots,t_n\}$ and visual tokens $\{v_1,\dots,v_m\}$.

The text tokens usually include system tokens, instruction tokens, and the output tokens generated by MLLMs.
The text tokens are embedded via the LLM embedding layer $E_{t}(\cdot)$ as:
\begin{equation}
e_{t_i} = E_t(t_i), \quad i \in [1, n].
\end{equation}
The visual tokens are extracted by a vision encoder ($E_{v}(\cdot)$), e.g., CLIP,  and projected to the LLM embedding space through a projection layer ($f_{mlp}(\cdot)$) as:
\begin{equation}
e_{v_i} = f_{mlp}(E_v(v_i)), \quad i \in [1,m].
\end{equation}
Finally, text tokens embeddings and visual tokens embeddings are interweaved to form a multimodal input sequence:
\begin{equation}
\scalebox{0.9}{$e_{x_i} = [ e_{t_1}, \cdots, e_{t_p}, e_{v_1}, \cdots, e_{v_m}, e_{t_{p+1}}, \cdots, e_{t_n} ], i \in [1, m + n],$}
\end{equation}
where $p$ is the number of system tokens, $n$ is the number of text tokens, and $m$ is the number of visual tokens.

The multimodal embeddings $e_{x_i}$ are then fed into the MLLM to generate the corresponding feature representations $h_{x_i}$, which are then processed by the MLLM's projection head to produce the final output token probabilities $p(x_{i+1}|x_{<i+1})$.
The next output token $x_{i+1}$ is generated by sampling from the output distribution $p(x_{i+1}|x_{<i+1})$ and is appended to the input sequence for the next decoding step.

LLaVA uses a two-stage training process: (1) freeze LLM and vision encoder to train the projection layer $f_{mlp}(\cdot)$ on image-text pairs; (2) jointly fine-tune the LLM and projection layer on a multimodal instruction dataset.

\subsection{Feature-level Speculative Decoding}
\label{sec:feature-level_speculative_decoding}
Li \emph{et al.}~\cite{li2024eagle} introduce feature-level speculative decoding to accelerate LLM generation via a two-phase process: \mbox{\textbf{drafting}~and~\textbf{verification}}.

\paragraph{Draft Phase}
A lightweight draft model $M_q$ predicts the next token's feature $h_{x_{i+1}}$, given the prefix hidden states $\{h_{x_0},\dots,h_{x_i}\}$ and input embeddings $\{e_{x_1},\dots,e_{x_{i+1}}\}$:
\begin{align}
    h_{x_{i+1}} &= M_q(f_{x_0},\dots,f_{x_i}), \nonumber\\
    f_{x_i} &= f_{\text{down}}(\text{Concat}(h_{x_i},e_{x_{i+1}})).
    \label{eq:concat}
\end{align}
Here, $\text{Concat}(\cdot)$ concatenates features in the last dimension, and $f_{\text{down}}(\cdot)$ projects them to match the dimension of $h_{x_i}$.
$f_{x_i}$ is fed into a transformer layer that constitutes the draft model $M_q$.
The predicted feature $h_{x_{i+1}}$ is passed through the target model's projection head to obtain the predicted token probability $q(\hat{x}_{i+1}|x_{<i+1})$ as:
\begin{equation}
    q(\hat{x}_{i+1}|x_{<i+1}) = f_{LM_{head}}(h_{x_{i+1}}).
\end{equation}
This process is repeated for $\gamma$ steps, generating a sequence of $\gamma$ candidate tokens $\hat{x}_{i+1:i+\gamma}$ and their associated probabilities $q(\hat{x}_{i+n}|x_{<t+n})$ for $n = 1,\cdots, \gamma$.
A tree-structured attention mechanism~\cite{li2024eagle} is employed in this phase. It builds a hierarchical draft tree with depth $m$, generating over $m$ tokens via $m$ forward passes. At each step, the model expands each leaf node by generating multiple candidate tokens in parallel, growing the tree breadth-wise.
\paragraph{Verification Phase}
The target model $M_p$ verifies the candidate tokens in parallel by computing their probabilities $p(\hat{x}_{t+n}|x_{<t+n})$. The acceptance probability for a candidate token $\hat{x}_{t+n}$ is defined as:
\begin{equation}
    p_{\text{accept}}(\hat{x}_{t+n})=\min\!\left(1,\frac{p(\hat{x}_{t+n}|x_{<t+n})}{q(\hat{x}_{t+n}|x_{<t+n})}\right).
\end{equation}
If $\hat{x}_{t+j}$ is rejected, later draft tokens are discarded and a new token is sampled from the adjusted distribution:
\begin{equation}
    norm\left(\max\left(0,p(\hat{x}_{t + j}|x_{<t + j}) - q(\hat{x}_{t + j}|x_{<t + j})\right)\right).
\end{equation}
Here, the function $norm(\cdot)$ normalizes the adjusted distribution to ensure that it sums to one. If all candidate tokens are accepted, an additional token $\hat{x}_{t + \gamma + 1}$ is sampled from the target model's prediction distribution $p(\hat{x}_{t+ \gamma + 1}|x_{<t+ \gamma + 1})$.

\subsection{Baseline Setting}
\label{sec:baseline}
In LLMs, feature-level speculative decoding attains a notably greater average acceptance length compared to token-level speculative decoding~\cite{li2024eagle}.
To validate whether this advantage holds in MLLMs, we conduct an ablation study on the LLaVA-1.6-7B model.
The results in Table~\ref{tab:Ablation_level} show that feature-level speculative decoding still achieves significantly higher average acceptance length than token-level speculative decoding in MLLMs.
However, the average acceptance length of state-of-the-art feature-level speculative decoding in LLMs has achieved 5, while the average acceptance length of feature-level speculative decoding in MLLMs is only 3.66.
This indicates that the feature-level speculative decoding specifically designed for LLMs fails to fully exploit their acceleration advantages in MLLMs.
Thus, we build on feature-level speculative decoding to reimagine speculative decoding in MLLMs.

\begin{figure*}[!h]
    \centering
    \includegraphics[width=\textwidth]{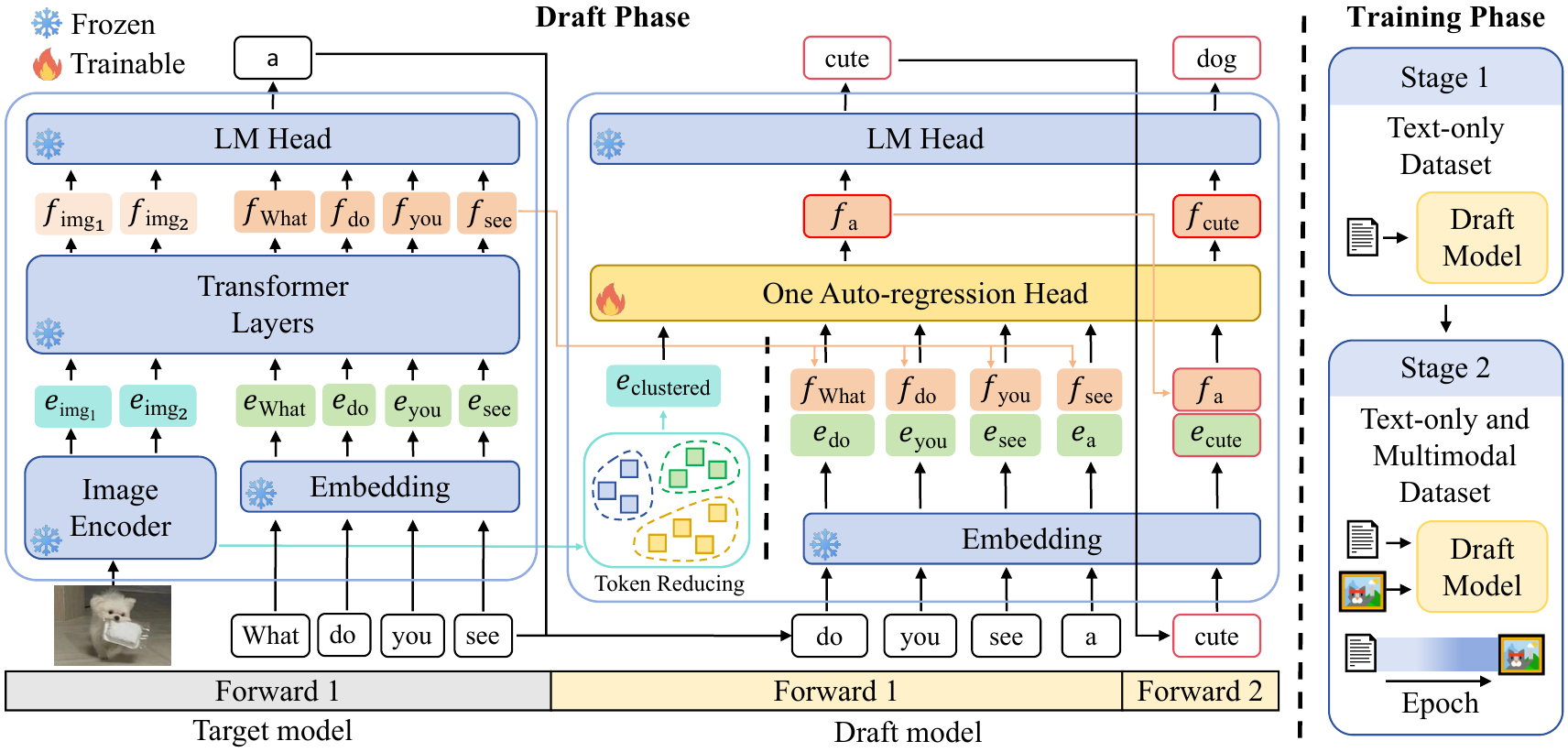}
    \caption{
    The framework of Multimodal Speculative Decoding (MSD). 
    The left illustrates the draft phase, while the right illustrates the training phase.
    \( e \) represents token embeddings, and \( f \) denotes features. 
    Red-bordered tokens are predicted by the draft model. 
    MSD concatenates text token hidden states with the next token embeddings, while directly inputting visual tokens embedding without concatenation during drafting, and applies a lightweight clustering-based module to compress the visual tokens seen by the draft model.
    MSD trains the draft model using a text-only instruction dataset in Stage 1, and then gradually introduces a multimodal instruction dataset in Stage 2.
    }
    \label{fig:framework}
\end{figure*}

\section{Multimodal Speculative Decoding}
\label{sec:msd}

In this section, we introduce Multimodal Speculative Decoding (MSD), a novel speculative decoding framework for MLLMs, as illustrated in Figure~\ref{fig:framework}.
%
%

\subsection{Decoupling Tokens of Different Modalities}
\label{sec:decoupling_tokens}

The most obvious difference between LLMs and MLLMs lies in the number of modalities.
The input to LLMs consists solely of the text modality, while the input to MLLMs not only includes textual modality but also incorporates additional modalities, such as the visual modality in the case of LLaVA.
A critical question arises: should input from different modalities be processed in a unified manner during drafting?
To answer this, we need to carefully examine the inherent characteristics and representational differences of the input from different modalities.

In text modality, the input is a sequence of words, which are tokenized and embedded into a sequence of text tokens.
In LLMs, each token's prediction is only dependent on the previous tokens due to the causal self-attention mechanism, showing a clear sequential dependency.
In contrast, in the visual modality, the input is images, which are divided into patches and then unfolded into a one-dimensional sequence of visual tokens.
These visual tokens are encoded by a visual encoder, e.g., CLIP, which uses bidirectional self-attention to process the image features.
Visual tokens are related to all visual tokens in the image, rather than just the previous tokens.
In traditional feature-level speculative decoding of LLMs~\cite{li2024eagle}, the hidden state of the current token is concatenated with the embedding of the next token to reduce the ambiguity of autoregressive generation, as formulated in Eq.\,(\ref{eq:concat}).
This design is valid for text tokens, as they have a clear sequential dependency.
However, for visual tokens, the displacement concatenation operation may hurt the information encoded in the visual tokens, as they do not have a sequential dependency like text tokens.
Thus, visual tokens should be processed differently from text tokens.

Based on the analysis of the inherent characteristics of different modal inputs, we decouple the processing of text and visual tokens in the draft model of MSD, as illustrated in Figure~\ref{fig:framework}.
Specifically, for text tokens, we keep the displacement concatenation operation in Eq.\,(\ref{eq:concat}), while for visual tokens, we instead directly adopt the visual token embeddings as inputs to the draft model.
The draft model predicts the next token's feature based on the concatenated text token features and the visual token embeddings as:
\begin{equation}
    h_{x_{i+1}} = M_q(f_{x_1}, \cdots, f_{x_i}), 
\end{equation}
where
\begin{equation}\label{eq:decouple}
    f_{x_i} = \begin{cases}
        f_{down}(\text{Concat}(h_{x_i},e_{x_{i+1}})), & \text{if } x_i \text{ is text token,} \\
        e_{x_i}, & \text{if } x_i \text{ is visual token.} \\
    \end{cases}
\end{equation}
Here, $\text{Concat}(\cdot)$ denotes the feature concatenation operation, and $h_{x_i}$ represents the corresponding hidden feature of $e_{x_i}$ in the target model. $f_{\text{down}}(\cdot)$ projects the concatenated features to match the dimension of $h_{x_i}$.
The remaining procedure is similar to that of the original feature-level speculative decoding method detailed in Section~\ref{sec:feature-level_speculative_decoding}.

As shown in Section~\ref{sec:input_ablation}, by decoupling the processing of text and visual tokens, MSD achieves a substantial performance improvement over the baseline.
This performance gain empirically validates our earlier analysis of the inherent characteristics of different modal inputs, highlighting the necessity of incorporating these characteristics into the design of the draft model for MLLMs.
We further provide a qualitative attention and draft-token analysis in Figure~\ref{fig:decoupling_attention}.

\subsection{Visual Token Reduction for the Draft Model}
We observe that the number of visual tokens fed into the draft model has a substantial impact on its computational cost.
As shown in Figure~\ref{fig:draft_token_prune}, reducing the visual token count markedly shortens the draft model’s forward latency, while the average acceptance length ($\tau$) remains nearly unchanged as long as some visual grounding is preserved.
This suggests that a small set of representative visual tokens is sufficient for the draft model of MLLMs, motivating the use of a heavily reduced visual input.

We therefore adopt a lightweight visual token reduction module for the draft model that performs a single pass of k-means style assignment and averaging over visual tokens to obtain a compact set of representatives~\cite{kmeans}.

Concretely, we apply visual token reduction on the visual token embeddings $\{e_{v_i}\}_{i=1}^{m}$ defined in Section~\ref{sec:decoupling_tokens}, where $m$ denotes the number of visual tokens fed into the target model.
We choose a small number of clusters $K \ll m$ and initialize $K$ cluster centers $\{c_k\}_{k=1}^{K}$ in the visual embedding space randomly.
Each visual token is then assigned to its nearest center in Euclidean distance:
\begin{equation}
    a_i = \arg\min_{k \in \{1,\dots,K\}} \big\| e_{v_i} - c_k \big\|_2,
\end{equation}
where $a_i$ is the cluster index of the $i$-th visual token.
Given the assignments $\{a_i\}_{i=1}^{m}$, we average the visual embeddings within each cluster $C_k = \{ i \mid a_i = k \}$:
\begin{equation}
    \bar{e}_{v_k} = \frac{1}{|C_k|} \sum_{i \in C_k} e_{v_i}.
\end{equation}
The $K$ averaged embeddings $\{\bar{e}_{v_k}\}_{k=1}^{K}$ form a compressed set of visual tokens that replaces the original $\{e_{v_i}\}_{i=1}^{m}$ in the multimodal input sequence for the draft model.
Although the target and draft models therefore operate on visual sequences of different physical lengths, this does not cause an accumulated position mismatch. During drafting, each generated text token is explicitly assigned the position ID of its corresponding token in the target model's uncompressed sequence, and the same position mapping is used during training. Once candidate tokens are accepted, the text positions of both models advance synchronously. Thus, no padding is required; otherwise, padding would negate the sequence-length reduction achieved by visual token compression.
This single-pass $k$-means style aggregation only requires a single assignment-and-averaging pass over the visual token embeddings, so its computational overhead is negligible compared to the transformer layers of the draft model. Yet, it substantially shortens the visual sequence and preserves sufficient visual information for speculative decoding.

\subsection{Two-Stage Training for the Draft Model}
\label{sec:two_stage_training}

The difference in training data between LLMs and MLLMs affects how the draft model should be trained.
LLMs are trained on large-scale text-only data, which equips them with strong language modeling capabilities.
In contrast, MLLMs are fine-tuned from a well-trained LLM on vision instruction datasets to acquire both language modeling and visual perception abilities.
For LLMs, the draft model is naturally trained on text-only datasets, as there is no other modality involved. The draft model in LLMs is typically fine-tuned on text-only instruction datasets. 
However, for the draft model in MLLMs, we need to answer two important practical questions: Which dataset(s) should be used to train the draft model? Is the visual instruction dataset alone sufficient?

\begin{table}[tbp]
    \centering
    \caption{\label{tab:dataset}
        Average acceptance length ($\tau$) of the draft model trained on different datasets across three multimodal tasks~\cite{masry2022chartqa,Singh2019textvqa,guan2024hallusionbench}. The evaluation uses LLaVA-1.5-7B as the target model.
    }   
    \resizebox{0.9\columnwidth}{!}{%
    \begin{tabular}{@{}lccccc@{}}
    \toprule
    \textbf{Method} & \textbf{ChartQA} & \textbf{TextVQA} &  \textbf{Hallusion} & \textbf{Average} \\ 
    \midrule
    Text-only instruction dataset & 3.26 & 3.62 & 3.50 & 3.46 \\
    Vision instruction dataset & 3.45 & 3.81 & 3.72 & 3.66 \\
    \bottomrule
    \end{tabular}
    }
\end{table}

To investigate this, we conduct an ablation study by training the draft model separately on a text-only and a vision instruction dataset and evaluating their performance on multimodal benchmarks.
As shown in Table~\ref{tab:dataset}, the draft model trained on the text-only instruction dataset yields an average acceptance length of 3.46, whereas the model trained on the vision instruction dataset yields a higher value of 3.66.
Although the draft model trained on the vision instruction dataset performs consistently better than the one trained on the text-only dataset, it still falls considerably short of the average acceptance length typically achieved by speculative decoding in standard LLMs.

\begin{figure}[t]
    \centering
    \includegraphics[width=0.47\textwidth]{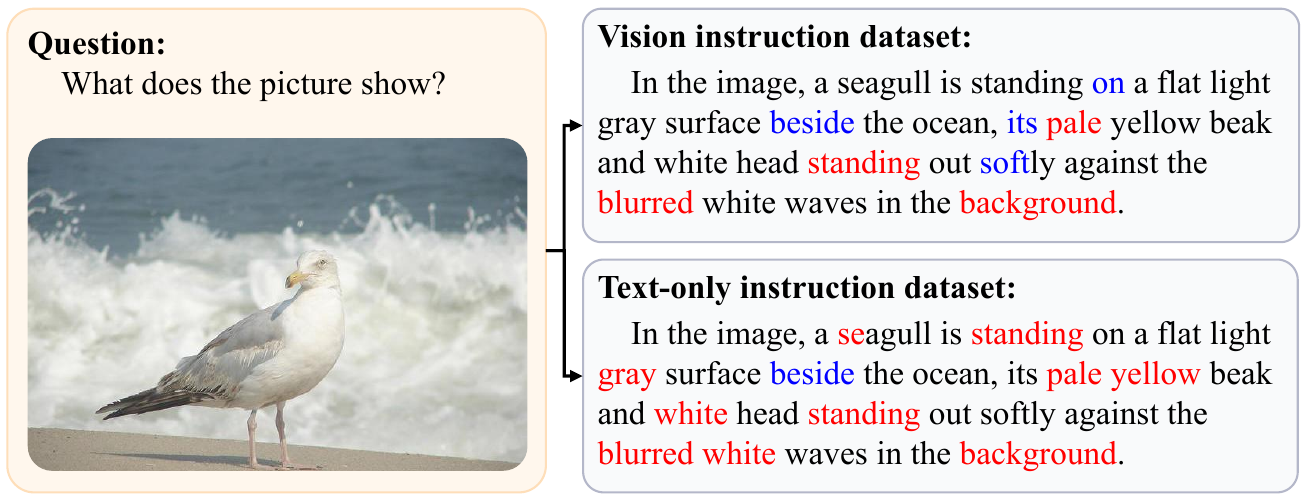}
    \caption{
    Comparison of models trained on text-only and vision instruction data, using LLaVA-1.6-7B as the target model for evaluation. Black tokens are judged as correct predictions, while red and blue tokens indicate vision-related errors and visually unrelated errors, respectively.
    \label{fig:vqa_example}
    }
\end{figure}

To dive deeper into the reasons behind this performance gap, we visualize the tokens that are rejected by the target model in Figure~\ref{fig:vqa_example}.
We find that the draft model trained on the vision instruction dataset tends to generate incorrect tokens that are unrelated to visual content, such as prepositions and conjunctions (e.g., ``on'' and ``its'').
While the draft model trained on the text-only instruction dataset is capable of generating these tokens, it fails to generate tokens that require visual perception, such as ``seagull'' and ``yellow''.
This indicates that the draft model trained on the vision instruction dataset is poor at language modeling, while the draft model trained on the text-only dataset is poor at visual perception.
Thus, it is not sufficient to train the draft model of MLLMs solely on the vision instruction dataset.

The draft model should therefore be trained on both text-only and vision instruction data to better balance language modeling capability and visual perception ability.
Based on this insight, we adopt a two-stage training scheme: (1) text-only language modeling and (2) multimodal training.

\paragraph{Stage 1: Fundamental Language Modeling Training}
In Stage 1, the draft model is trained on a text-only instruction dataset $D_{\text{text}}$ from ShareGPT~\cite{ShareGPT}.
It builds the draft model's foundational language modeling abilities, minimizing the likelihood of producing wrong tokens unrelated to visual content during the drafting stage.

\paragraph{Stage 2: Multimodal Fine-tuning}

In Stage 2, a natural approach is to directly fine-tune the draft model obtained from Stage 1 on a visual instruction dataset $D_{\text{visual}}$, aiming to endow the draft model with visual perception capabilities.
While this design follows a common two-stage fine-tuning setup, we find that such a direct dataset switch introduces a non-trivial side effect.
To assess whether the model still retains strong language modeling ability after visual fine-tuning, we evaluate the draft model on a diverse set of representative text tasks.
As shown in Table~\ref{tab:degradation}, directly fine-tuning on the visual dataset leads to a significant drop in the average acceptance length, with a reduction of 0.84 tokens on GSM8K~\cite{cobbe2021gsm8k}. This result reveals that a direct shift in training modality impairs the model’s language modeling ability.

\paragraph{Progressive Multimodal Data Mix Training}
To effectively mitigate this performance degradation, we propose a progressive multimodal data-mixing training strategy for Stage 2 of the overall training framework.
Instead of an abrupt shift to the visual dataset, we gradually increase its proportion over training epochs.
Specifically, at each epoch $t$ out of a total of $T$ epochs, the mixed dataset $D^t_{\text{mix}}$ is defined as:
\begin{equation}\label{eq:progressive_data}
    D^t_{\text{mix}} = D_{\text{text}} \cdot \frac{T - t}{T} + D_{\text{visual}} \cdot \frac{t}{T}.
\end{equation}
This progressive data scheduling strategy enables the draft model to gradually learn visual representations without compromising its core language-modeling capability.
As shown in Table~\ref{tab:degradation}, evaluated on text benchmarks (MT-Bench~\cite{zheng2023mtbench}, GSM8K~\cite{cobbe2021gsm8k}, and Natural Questions~\cite{Tom2019qa}), our progressive strategy effectively preserves the draft model's language modeling ability, incurring only a 0.06-token decrease in the average acceptance length on Natural Questions~\cite{Tom2019qa}. 
Further detailed ablations and comprehensive analysis are provided in Appendix~\ref{ablation: training strategy} for completeness.

\begin{table}[tbp]
    \centering
    \caption{\label{tab:degradation}
        Impact of draft-model training strategies on average acceptance length ($\tau$) across three text benchmarks~\cite{zheng2023mtbench,cobbe2021gsm8k,Tom2019qa}, evaluated with LLaVA-1.6-7B as the target model.
    }
    \resizebox{0.8\columnwidth}{!}{%
    \begin{tabular}{lcccc}
    \toprule
    \textbf{Draft Model Training} & \textbf{MT-Bench} & \textbf{GSM8K} & \textbf{Natural Ques.} & \textbf{Average} \\
    \midrule
    Stage 1 (Text-only) & 5.25 & 4.69 & 3.78 & 4.57 \\
    + Direct visual fine-tuning & 4.45 (-0.80) & 3.85 (-0.84) & 3.17 (-0.61) & 3.82 (-0.75) \\
    + Progressive visual fine-tuning & 5.10 (-0.15) & 4.48 (-0.21) & 3.72 (-0.06) & 4.43 (-0.14) \\
    \bottomrule
    \end{tabular}
    }
\end{table}

\section{Experiments} 
\label{sec:experi}

\begin{table*}[!t]
  \centering
  \caption{Speedup ratios $SR$ and average acceptance lengths $\tau$ of different speculative decoding methods on VQAv2~\cite{goyal2017vqav2}, AI2D~\cite{kembhavi2016ai2d}, SQA\_Image~\cite{lu2022sqa}, ChartQA~\cite{masry2022chartqa}, TextVQA~\cite{Singh2019textvqa}, Hallusion~\cite{guan2024hallusionbench}, MMB\_EN~\cite{Liu2023mmbench}, and MME~\cite{fu2023mme} with temperature $T \in \{0, 1\}$ using LLaVA-1.5-7B as the target model.}
  \resizebox{\linewidth}{!}{
    \begin{tabular}{cccccccccccccccccc}
    \toprule
          &       & \multicolumn{2}{c}{VQAv2} & \multicolumn{2}{c}{AI2D} & \multicolumn{2}{c}{SQA\_Image} & \multicolumn{2}{c}{ChartQA} & \multicolumn{2}{c}{TextVQA} & \multicolumn{2}{c}{Hallusion} & \multicolumn{2}{c}{MMB\_{EN}} & \multicolumn{2}{c}{MME} \\
    \midrule
    Model & Method & $SR$ & $\tau$ & $SR$ & $\tau$ & $SR$ & $\tau$ & $SR$ & $\tau$ & $SR$ & $\tau$ & $SR$ & $\tau$ & $SR$ & $\tau$ & $SR$ & $\tau$ \\
    \midrule
    \multicolumn{18}{c}{Temperature = 0} \\
    \midrule
    \multirow{6}{*}{LLaVA-1.5-7B} 
      & Baseline   & 2.45$\times$ & 4.23 & 2.23$\times$ & 3.46 & 2.18$\times$ & 3.47 & 2.21$\times$ & 3.45 & 1.99$\times$ & 3.81 & 2.36$\times$ & 3.72 & 2.28$\times$ & 3.58 & 2.37$\times$ & 3.67 \\
      & Lookahead  & 1.28$\times$ & 1.33 & 1.37$\times$ & 1.45 & 1.35$\times$ & 1.45 & 1.32$\times$ & 1.42 & 1.22$\times$ & 1.31 & 1.41$\times$ & 1.50 & 1.32$\times$ & 1.41 & 1.34$\times$ & 1.40 \\
      & Medusa     & 1.32$\times$ & 2.28 & 1.39$\times$ & 1.79 & 1.34$\times$ & 1.79 & 1.36$\times$ & 1.80 & 1.22$\times$ & 1.89 & 1.26$\times$ & 1.91 & 1.40$\times$ & 1.86 & 1.28$\times$ & 1.83 \\
      & Text\_Only & 2.23$\times$ & 4.01 & 1.98$\times$ & 3.14 & 1.94$\times$ & 3.30 & 2.02$\times$ & 3.21 & 1.65$\times$ & 3.55 & 2.11$\times$ & 3.45 & 2.01$\times$ & 3.29 & 2.18$\times$ & 3.34 \\
      & ViSpec     & 2.75$\times$ & 4.62 & 2.62$\times$ & 4.25 & 2.50$\times$ & 4.37 & 2.80$\times$ & 4.33 & 2.38$\times$ & 4.34 & 2.88$\times$ & 4.70 & 2.74$\times$ & 4.55 & 2.70$\times$ & 4.42 \\
      & \textbf{MSD}        & \textbf{2.86$\times$} & \textbf{5.01} & \textbf{2.70$\times$} & \textbf{4.48} & \textbf{2.66$\times$} & \textbf{4.61} & \textbf{2.94$\times$} & \textbf{4.52} & \textbf{2.52$\times$} & \textbf{4.52} & \textbf{3.03$\times$} & \textbf{4.90} & \textbf{2.83$\times$} & \textbf{4.82} & \textbf{2.79$\times$} & \textbf{4.60} \\
    \midrule
    \multicolumn{18}{c}{Temperature = 1} \\
    \midrule
    \multirow{6}{*}{LLaVA-1.5-7B} 
      & Baseline & 1.64$\times$ & 3.08 & 1.50$\times$ & 2.61 & 1.74$\times$ & 2.58 & 1.62$\times$ & 2.53 & 1.78$\times$ & 2.79 & 1.66$\times$ & 2.85 & 1.68$\times$ & 2.79 & 1.61$\times$ & 2.64 \\
      & Lookahead  & 1.19$\times$ & 1.29 & 1.26$\times$ & 1.34 & 1.24$\times$ & 1.35 & 1.23$\times$ & 1.33 & 1.18$\times$ & 1.26 & 1.34$\times$ & 1.42 & 1.26$\times$ & 1.34 & 1.20$\times$ & 1.29 \\
      & Medusa     & 1.44$\times$ & 2.38 & 1.48$\times$ & 1.91 & 1.42$\times$ & 1.89 & 1.47$\times$ & 1.89 & 1.34$\times$ & 2.00 & 1.45$\times$ & 1.98 & 1.50$\times$ & 1.95 & 1.38$\times$ & 1.93 \\
      & Text\_Only & 1.48$\times$ & 2.89 & 1.33$\times$ & 2.45 & 1.49$\times$ & 2.50 & 1.32$\times$ & 2.21 & 1.49$\times$ & 2.41 & 1.37$\times$ & 2.50 & 1.43$\times$ & 2.67 & 1.33$\times$ & 2.46 \\
      & ViSpec   & 2.00$\times$ & 3.38 & 2.00$\times$ & 3.12 & 2.04$\times$ & 3.17 & 2.03$\times$ & 3.07 & 2.08$\times$ & 3.10 & 2.05$\times$ & 3.37 & 2.07$\times$ & 3.36 & 1.94$\times$ & 3.08 \\
      & \textbf{MSD}      & \textbf{2.10$\times$} & \textbf{3.55} & \textbf{2.04$\times$} & \textbf{3.22} & \textbf{2.19$\times$} & \textbf{3.38} & \textbf{2.11$\times$} & \textbf{3.18} & \textbf{2.17$\times$} & \textbf{3.20} & \textbf{2.16$\times$} & \textbf{3.54} & \textbf{2.29$\times$} & \textbf{3.52} & \textbf{2.00$\times$} & \textbf{3.15} \\
    \bottomrule
    \end{tabular}
  }
  \label{tab:results_llava15}
\end{table*}

\begin{table*}[!t]
  \centering
  \caption{Speedup ratios $SR$ and average acceptance lengths $\tau$ on the same benchmarks with temperature $T \in \{0, 1\}$ using LLaVA-1.6-7B, LLaVA-1.6-13B~\cite{liu2024llavanext}, and Qwen2.5-VL-7B~\cite{bai2025qwen25vl} as target models.}
  \resizebox{\linewidth}{!}{
    \begin{tabular}{cccccccccccccccccc}
    \toprule
          &       & \multicolumn{2}{c}{VQAv2} & \multicolumn{2}{c}{AI2D} & \multicolumn{2}{c}{SQA\_Image} & \multicolumn{2}{c}{ChartQA} & \multicolumn{2}{c}{TextVQA} & \multicolumn{2}{c}{Hallusion} & \multicolumn{2}{c}{MMB\_{EN}} & \multicolumn{2}{c}{MME} \\
    \midrule
    Model & Method & $SR$ & $\tau$ & $SR$ & $\tau$ & $SR$ & $\tau$ & $SR$ & $\tau$ & $SR$ & $\tau$ & $SR$ & $\tau$ & $SR$ & $\tau$ & $SR$ & $\tau$ \\
    \midrule
    \multicolumn{18}{c}{Temperature = 0} \\
    \midrule
    \multirow{3}{*}{LLaVA-1.6-7B} 
      & Baseline & 1.70$\times$ & 3.82 & 2.03$\times$ & 3.54 & 2.01$\times$ & 3.65 & 2.11$\times$ & 3.52 & 1.71$\times$ & 3.99 & 2.23$\times$ & 3.86 & 2.08$\times$ & 3.79 & 2.14$\times$ & 3.69 \\
      & ViSpec   & 2.12$\times$ & 4.45 & 2.64$\times$ & 4.22 & 2.44$\times$ & 4.25 & 2.56$\times$ & 4.29 & 2.16$\times$ & 4.25 & 2.81$\times$ & 4.70 & 2.77$\times$ & 4.56 & 2.66$\times$ & 4.46 \\
      & \textbf{MSD}      & \textbf{2.21$\times$} & \textbf{4.67} & \textbf{2.79$\times$} & \textbf{4.41} & \textbf{2.59$\times$} & \textbf{4.41} & \textbf{2.72$\times$} & \textbf{4.49} & \textbf{2.27$\times$} & \textbf{4.39} & \textbf{2.91$\times$} & \textbf{4.79} & \textbf{2.81$\times$} & \textbf{4.66} & \textbf{2.81$\times$} & \textbf{4.67} \\
    \midrule
    \multirow{3}{*}{LLaVA-1.6-13B} 
      & Baseline & 2.04$\times$ & 4.22 & 2.49$\times$ & 3.79 & 2.50$\times$ & 3.86 & 2.51$\times$ & 3.94 & 2.00$\times$ & 4.14 & 2.58$\times$ & 4.14 & 2.38$\times$ & 3.90 & 2.66$\times$ & 3.86 \\
      & ViSpec   & 2.30$\times$ & 4.51 & 2.76$\times$ & 4.29 & 2.90$\times$ & 4.46 & 2.72$\times$ & 4.39 & 2.27$\times$ & 4.42 & 2.85$\times$ & 4.68 & 2.81$\times$ & 4.43 & 3.10$\times$ & 4.59 \\
      & \textbf{MSD}      & \textbf{2.40$\times$} & \textbf{4.83} & \textbf{2.97$\times$} & \textbf{4.54} & \textbf{3.02$\times$} & \textbf{4.64} & \textbf{2.95$\times$} & \textbf{4.62} & \textbf{2.38$\times$} & \textbf{4.61} & \textbf{3.10$\times$} & \textbf{5.01} & \textbf{2.94$\times$} & \textbf{4.65} & \textbf{3.24$\times$} & \textbf{4.83} \\
    \midrule
    \multirow{3}{*}{Qwen2.5-VL-7B} 
      & Baseline & 2.12$\times$ & 3.54 & 1.72$\times$ & 2.85 & 1.66$\times$ & 2.80 & 1.81$\times$ & 3.09 & 1.72$\times$ & 3.05 & 1.69$\times$ & 3.08 & 1.76$\times$ & 2.90 & 1.65$\times$ & 2.91 \\
      & ViSpec   & 2.39$\times$ & 3.67 & 2.10$\times$ & 3.68 & 2.20$\times$ & 3.64 & 2.50$\times$ & 4.04 & 2.03$\times$ & 3.58 & 2.07$\times$ & 4.03 & 2.34$\times$ & 3.64 & 2.13$\times$ & 3.70 \\
      & \textbf{MSD}      & \textbf{2.61$\times$} & \textbf{4.36} & \textbf{2.41$\times$} & \textbf{4.21} & \textbf{2.43$\times$} & \textbf{4.12} & \textbf{2.60$\times$} & \textbf{4.54} & \textbf{2.21$\times$} & \textbf{4.19} & \textbf{2.32$\times$} & \textbf{4.54} & \textbf{2.60$\times$} & \textbf{4.18} & \textbf{2.41$\times$} & \textbf{4.33} \\
    \midrule
    \multicolumn{18}{c}{Temperature = 1} \\
    \midrule
    \multirow{3}{*}{LLaVA-1.6-7B} 
      & Baseline & 1.65$\times$ & 3.13 & 1.70$\times$ & 2.75 & 1.77$\times$ & 2.85 & 1.76$\times$ & 2.75 & 1.65$\times$ & 2.86 & 1.96$\times$ & 3.02 & 1.50$\times$ & 2.93 & 1.61$\times$ & 2.70 \\
      & ViSpec   & 1.93$\times$ & 3.54 & 2.01$\times$ & 3.14 & 2.13$\times$ & 3.22 & 1.99$\times$ & 3.28 & 2.01$\times$ & 3.13 & 2.11$\times$ & 3.49 & 1.71$\times$ & 3.31 & 2.07$\times$ & 3.28 \\
      & \textbf{MSD}      & \textbf{1.97$\times$} & \textbf{3.56} & \textbf{2.06$\times$} & \textbf{3.16} & \textbf{2.20$\times$} & \textbf{3.30} & \textbf{2.09$\times$} & \textbf{3.34} & \textbf{2.09$\times$} & \textbf{3.20} & \textbf{2.18$\times$} & \textbf{3.56} & \textbf{1.80$\times$} & \textbf{3.38} & \textbf{2.08$\times$} & \textbf{3.30} \\
    \midrule
    \multirow{3}{*}{LLaVA-1.6-13B} 
      & Baseline & 1.78$\times$ & 3.14 & 1.51$\times$ & 2.81 & 1.64$\times$ & 2.95 & 1.95$\times$ & 3.01 & 1.59$\times$ & 2.99 & 2.06$\times$ & 3.26 & 1.52$\times$ & 3.03 & 2.13$\times$ & 2.88 \\
      & ViSpec   & 2.06$\times$ & 3.45 & 1.76$\times$ & 3.16 & 1.75$\times$ & 3.12 & 2.16$\times$ & 3.45 & 1.84$\times$ & 3.29 & 2.25$\times$ & 3.54 & 1.94$\times$ & 3.37 & 2.37$\times$ & 3.32 \\
      & \textbf{MSD}      & \textbf{2.15$\times$} & \textbf{3.56} & \textbf{1.89$\times$} & \textbf{3.31} & \textbf{1.90$\times$} & \textbf{3.32} & \textbf{2.27$\times$} & \textbf{3.65} & \textbf{1.96$\times$} & \textbf{3.34} & \textbf{2.32$\times$} & \textbf{3.69} & \textbf{2.01$\times$} & \textbf{3.48} & \textbf{2.49$\times$} & \textbf{3.49} \\
    \midrule
    \multirow{3}{*}{Qwen2.5-VL-7B} 
      & Baseline & 2.08$\times$ & 3.54 & 1.70$\times$ & 2.85 & 1.67$\times$ & 2.80 & 1.87$\times$ & 3.09 & 1.60$\times$ & 3.05 & 1.72$\times$ & 3.08 & 1.77$\times$ & 2.90 & 1.75$\times$ & 2.91 \\
      & ViSpec   & 1.86$\times$ & 3.25 & 1.93$\times$ & 3.31 & 1.78$\times$ & 3.32 & 2.20$\times$ & 3.82 & 1.63$\times$ & 3.27 & 1.87$\times$ & 3.58 & 1.91$\times$ & 3.34 & 1.87$\times$ & 3.26 \\
      & \textbf{MSD}      & \textbf{2.26$\times$} & \textbf{4.18} & \textbf{2.29$\times$} & \textbf{3.91} & \textbf{2.06$\times$} & \textbf{3.85} & \textbf{2.31$\times$} & \textbf{3.94} & \textbf{1.96$\times$} & \textbf{3.54} & \textbf{2.10$\times$} & \textbf{3.82} & \textbf{2.16$\times$} & \textbf{3.72} & \textbf{2.11$\times$} & \textbf{3.99} \\
    \bottomrule
    \end{tabular}
  }
  \label{tab:results_generalization}
\end{table*}

\subsection{Model Setup and Benchmarks}
%

%
MSD is applied to three MLLMs: LLaVA-1.5~\cite{liu2023llava1.5}, LLaVA-1.6~\cite{liu2024llavanext}, and Qwen2.5VL~\cite{bai2025qwen25vl}. The draft model, which shares hyperparameters with the target model, is a lightweight single-layer Transformer used for feature-level prediction. More detailed benchmarks and training details are provided in Appendix~\ref{app:train_detail}.

\subsection{Metrics}
We assess the performance of our MSD using the following widely used metrics in speculative decoding:

\paragraph{Average Acceptance Length \(\tau\)} This metric measures the average number of tokens accepted by the target model in each drafting-verification cycle. It reflects the effectiveness of the draft model in predicting tokens that align with the target model's output. A higher \(\tau\) indicates better draft accuracy and fewer verification rounds required.

\paragraph{Speedup Ratio $SR$} The $SR$ measures the efficiency of speculative decoding over vanilla autoregressive decoding. It depends on factors such as acceptance rate, draft length, prefilling time, and output length.
Following prior work~\cite{chen2024magicdec}, we include the derivation in Appendix~\ref{app:speedup}. Briefly, $SR$ grows with longer outputs and shorter prefilling time.

\subsection{Main Results}
We compare MSD with several speculative decoding methods, including
Baseline, Lookahead~\cite{fu2024lade}, Medusa~\cite{cai2024medusa},
Text\_Only~\cite{gagrani2024spd_mllm}, and ViSpec~\cite{vispec}.
Detailed implementation settings and the exact draft model configurations adopted for each method are provided in Appendix~\ref{app:baselines} for reproducibility.

Table~\ref{tab:results_llava15} and Table~\ref{tab:results_generalization} report the average acceptance length and corresponding speedup ratio across different tasks, temperature settings, and model scales, clearly showing that our method consistently achieves robust and state-of-the-art inference speedups.

From Table~\ref{tab:results_llava15}, MSD consistently yields larger acceptance lengths $\tau$ and higher speedup ratios $SR$ than all other baselines on LLaVA-1.5-7B across datasets and temperatures.
LLM SD methods such as Lookahead and Medusa bring only modest gains when extended to multimodal decoding, and the Text\_Only further shows that ignoring visual features consistently degrades performance on these multimodal benchmarks.
Compared with ViSpec~\cite{vispec}, which compresses all visual tokens into a single global feature, MSD decouples text and vision processing, retains richer visual context via lightweight token reduction, and adopts a capability-oriented training strategy, leading to stronger practical speedups on multimodal benchmarks.
Table~\ref{tab:results_generalization} further demonstrates that these advantages remain robust and transferable across different backbones, including LLaVA-1.6-7B, LLaVA-1.6-13B~\cite{liu2024llavanext}, and Qwen2.5-VL-7B~\cite{bai2025qwen25vl}.
We further verify MSD's downstream accuracy under greedy decoding in Table~\ref{tab:lossless_verification}.

\raggedbottom
Although MSD matches LLM speculative decoding in average acceptance length, its overall speedup ratio remains lower, mainly because MLLM benchmarks are usually characterized by longer input contexts and shorter output responses than LLM benchmarks, which fundamentally limits achievable acceleration; a formal analysis is provided in Appendix~\ref{appendix:speedup_gap}.

\textit{Latency Breakdown.} We further profile Baseline and MSD on 100 AI2D samples using LLaVA-1.6-7B at $T=0$, batch size 1, and an NVIDIA A100 GPU. GPU components are timed using CUDA Events.

\begin{table*}[t]
    \centering
    
    \caption{System-level latency breakdown on AI2D. Draft is the first draft forward pass, Verify is measured per round, Rounds and $\tau$ denote the average verification-round count and acceptance length, respectively, and Total is the complete decoding latency per sample. All latency values are in ms.}
    \label{tab:latency_breakdown}
    \resizebox{0.70\textwidth}{!}{%
    \begin{tabular}{@{}lccccccccc@{}}
        \toprule
        \multirow{2}{*}{\textbf{Method}} & \multicolumn{3}{c}{\textbf{Common Stages}} & \multicolumn{5}{c}{\textbf{Drafting \& Verification}} & \multirow{2}{*}{\textbf{Total}} \\
        \cmidrule(lr){2-4}\cmidrule(lr){5-9}
        & \textbf{Vision} & \textbf{Projector} & \textbf{Prefill} & \textbf{Reduction} & \textbf{Draft} & \textbf{Verify} & \textbf{Rounds} & $\boldsymbol{\tau}$ & \\
        \midrule
        LLaVA-1.6-7B & 16.10 & 0.67 & 308.39 & -- & -- & -- & -- & -- & 6416.34 \\
        + Baseline & 15.89 & 0.66 & 308.43 & -- & 25.88 & 28.81 & 62.75 & 3.54 & 2806.74 \\
        + MSD & 15.93 & 0.66 & 307.94 & 1.28 & 9.16 & 28.59 & 48.54 & 4.41 & 2266.73 \\
        \bottomrule
    \end{tabular}%
    }
\end{table*}

As shown in Table~\ref{tab:latency_breakdown}, Baseline and MSD have nearly identical vision-encoder, projector, target-prefill, and per-round verification costs. MSD benefits from a shorter first draft forward pass and a higher average acceptance length, which reduces the average number of verification rounds from 62.75 to 48.54. Consequently, the complete decoding latency decreases from 2806.74 ms to 2266.73 ms per sample.

\begin{table}[t]
    \centering
    \caption{
      Ablation study on the contributions of MSD components, evaluated by average acceptance length.
    }
    \label{tab:ablation_individual}
    \resizebox{\columnwidth}{!}{%
    \begin{tabular}{@{}lccccc@{}}
    \toprule
    \textbf{Model}
    & \textbf{ChartQA} $\uparrow$
    & \textbf{AI2D} $\uparrow$
    & \textbf{MMB} $\uparrow$
    & \textbf{Hallusion} $\uparrow$
    & \textbf{Average} \\
    \midrule
    Baseline
    & 3.52
    & 3.54
    & 3.79
    & 3.86
    & 3.68 \\
    \midrule
    + Input Tokens Decoupling
    & 3.93 \textbf{(+0.41)}
    & 3.96 \textbf{(+0.42)}
    & 4.25 \textbf{(+0.46)}
    & 4.33 \textbf{(+0.47)}
    & 4.12 \textbf{(+0.44)} \\

    + Two-Stage (Direct)
    & 4.16 \textbf{(+0.64)}
    & 4.05 \textbf{(+0.51)}
    & 4.32 \textbf{(+0.53)}
    & 4.42 \textbf{(+0.56)}
    & 4.24 \textbf{(+0.56)} \\

    + Two-Stage (Progressive)
    & 4.27 \textbf{(+0.75)}
    & 4.24 \textbf{(+0.70)}
    & 4.51 \textbf{(+0.72)}
    & 4.56 \textbf{(+0.70)}
    & 4.40 \textbf{(+0.72)} \\
    \midrule
    MSD
    & 4.49 \textbf{(+0.97)}
    & 4.41 \textbf{(+0.87)}
    & 4.66 \textbf{(+0.87)}
    & 4.79 \textbf{(+0.93)}
    & 4.59 \textbf{(+0.91)} \\
    \bottomrule
    \end{tabular}
    }
\end{table}

\subsection{Ablation study}
\label{sec:input_ablation}
In this section, we conduct an ablation study to assess the impact of MSD’s key components, including the input token decoupling strategy, the visual token reduction module, and the progressive two-stage training strategy, along with their respective variants.
All experiments are conducted on LLaVA-1.6-7B with a temperature of 0.

\paragraph{Ablation on Key Modules of MSD}
As shown in Table~\ref{tab:ablation_individual}, both the input tokens decoupling strategy and progressive two-stage training contribute significantly to MSD's performance gains (the effect of the token reducing module is discussed in the following section).
The decoupling strategy increases the average acceptance length by about +0.44, underscoring the benefit of handling visual and textual tokens separately in the draft model.
This design is particularly advantageous for tasks requiring fine-grained visual reasoning, such as ChartQA~\cite{masry2022chartqa}.
The improvement brought by two-stage training further confirms that both language modeling and visual perception are essential capabilities for the draft model.
In particular, our progressive two-stage training yields a larger gain (approximately +0.72 vs +0.56), as the gradual introduction of visual data enables a smoother transition across modalities.
This mitigates performance degradation caused by abrupt modality shifts and preserves the model’s core language modeling ability.
By combining these components, MSD achieves the highest average acceptance length across all tasks, outperforming the Baseline by an average of 0.91, indicating that the proposed improvements are complementary.

\begin{figure}[t]
    \centering
    \includegraphics[width=0.5\textwidth]{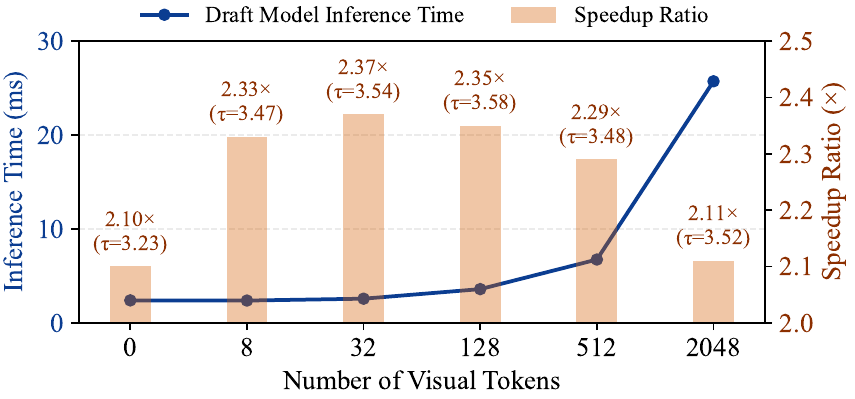}
    \caption{
    Impact of visual token reduction on the draft model's forward latency and the overall inference speedup, evaluated on the ChartQA benchmark~\cite{masry2022chartqa}.
    \label{fig:draft_token_prune}
    }
\end{figure}

\paragraph{Ablation on Draft Model Visual Tokens.}
As shown in Figure~\ref{fig:draft_token_prune}, progressively reducing the number of visual tokens significantly shortens the draft model’s forward latency, while the average acceptance length ($\tau$) remains largely stable within a reasonable compression range. This suggests that visual inputs contain substantial redundancy for drafting purposes, and a compact set of representative tokens is sufficient to maintain stable speculative verification. However, completely removing visual tokens ($K{=}0$) substantially reduces $\tau$, indicating that some degree of visual grounding is necessary for reliable draft predictions. Balancing computational efficiency and acceptance stability, we adopt $K=32$ visual tokens as the default configuration.

\paragraph{Comparison with Alternative Visual Token Reduction Methods}
We compare K-Means with representative alternatives under the same $K=32$ budget, draft architecture, training data, and evaluation protocol. K-Means groups visual features and replaces each cluster with its centroid without trainable parameters. \textbf{FasterVLM}~\cite{zhang2024fastervlm} ranks patches using \texttt{[CLS]}-to-patch attention and discards low-scoring tokens. \textbf{ToMe}~\cite{bolya2023tome} repeatedly matches and averages similar features at the draft-model input until reaching the token budget. \textbf{Resampler}~\cite{alayrac2022flamingo} aggregates visual features with learned cross-attention queries, adding about 21M parameters.

\begin{table}[t]
  \centering
  
  \caption{Comparison of visual token reduction methods on LLaVA-1.6-7B at $T=0$. All training and inference are conducted on an NVIDIA A100 GPU. Each entry reports $\tau$ / end-to-end speedup.}
  \label{tab:token_reduction_comparison}
  \resizebox{\columnwidth}{!}{%
  \begin{tabular}{@{}lccccc@{}}
  \toprule
  \textbf{Method} & \textbf{ChartQA} & \textbf{AI2D} & \textbf{MMB$_{\rm EN}$} & \textbf{Hallusion} & \textbf{Average} \\
  \midrule
  K-Means & 4.46 / \textbf{2.58$\times$} & 4.40 / \textbf{2.91$\times$} & 4.59 / \textbf{2.95$\times$} & 4.90 / \textbf{2.93$\times$} & 4.59 / \textbf{2.84$\times$} \\
  FasterVLM & 4.40 / 2.53$\times$ & 4.35 / 2.84$\times$ & 4.51 / 2.87$\times$ & 4.80 / 2.86$\times$ & 4.52 / 2.78$\times$ \\
  ToMe & 4.45 / 2.48$\times$ & 4.32 / 2.77$\times$ & 4.54 / 2.80$\times$ & 4.81 / 2.79$\times$ & 4.53 / 2.71$\times$ \\
  Resampler & \textbf{4.48} / 2.42$\times$ & \textbf{4.44} / 2.69$\times$ & \textbf{4.63} / 2.72$\times$ & \textbf{4.92} / 2.70$\times$ & \textbf{4.62} / 2.63$\times$ \\
  \bottomrule
  \end{tabular}%
  }
\end{table}

K-Means achieves the highest average end-to-end speedup of 2.84$\times$ as shown in Table~\ref{tab:token_reduction_comparison}.
\par\vspace{0.3\baselineskip}

\begin{table}[t]
  \centering
  \caption{
    Ablation study on different strategies for handling visual tokens, evaluated in terms of the average acceptance length across validation benchmarks.
  }
  \label{tab:decouple_input}
  \resizebox{\columnwidth}{!}{%
  \begin{tabular}{@{}lcccc@{}}
  \toprule
  \textbf{Method} & \textbf{ChartQA} & \textbf{AI2D} & \textbf{MMB} & \textbf{Hallusion} \\ 
  \midrule
  Baseline    & 3.52 & 3.54 & 3.79 & 3.86 \\
  Visual Feature & 3.91 (\textbf{+0.39}) & 3.92 (\textbf{+0.38}) & 4.22 (\textbf{+0.43}) & 4.30 (\textbf{+0.44}) \\
  Visual Embedding & 3.93 (\textbf{+0.41}) & 3.96 (\textbf{+0.42}) & 4.25 (\textbf{+0.46}) & 4.33 (\textbf{+0.47}) \\
  \bottomrule
  \end{tabular}
  }
  \footnotesize
\end{table}

\paragraph{Ablation on Input Token Decoupling Strategy}

To validate our input-token decoupling, we ablate three ways of handling visual tokens:
(i) Baseline, which concatenates visual tokens with the next-token hidden state (Section~\ref{sec:baseline});
(ii) Visual Feature, which uses LLM hidden features without concatenation;
and (iii) Visual Embedding, which uses the original visual embeddings from the target model’s input.
As shown in Table~\ref{tab:decouple_input}, the non-decoupled Baseline underperforms both decoupled variants, confirming the benefit of separating visual from textual tokens in the draft model.
Between the two decoupled methods, Visual Embedding yields a slightly higher average acceptance length, further suggesting that the original encoder embeddings preserve cleaner and more faithful visual information compared to LLM hidden features.

\begin{figure}[t]
    \centering
    \includegraphics[width=\columnwidth]{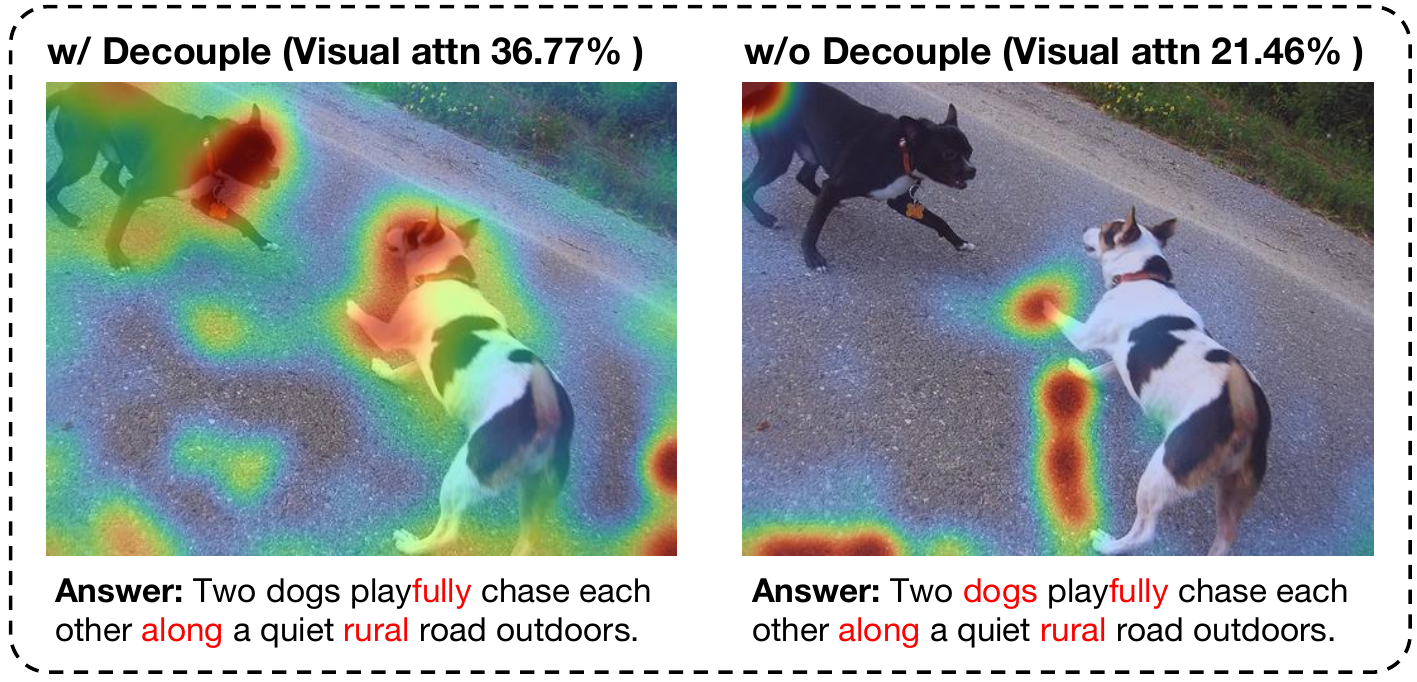}
    
    \caption{Single-layer draft-model attention for the first output token and draft-token comparison with and without input token decoupling on LLaVA-1.6-7B. The prompt is ``Provide a one-sentence caption for the provided image.'' Percentages show visual-token attention; red words mark draft tokens rejected by the target model.}
    \label{fig:decoupling_attention}
\end{figure}

\paragraph{Qualitative Analysis of Input Token Decoupling.}
Figure~\ref{fig:decoupling_attention} compares the draft model with and without input token decoupling under the same input. Since our draft model contains a single Transformer layer, we visualize this layer's attention distribution over the input sequence when predicting the first output token. The percentage in the figure denotes the total attention assigned to visual tokens. With decoupling, the attention covers both dogs, whereas shifted concatenation focuses more on partial or background regions and reduces the visual-attention share from 36.77\% to 21.46\%. This weaker visual grounding is also reflected in candidate generation: the visually grounded token ``dogs'' is rejected without decoupling but accepted after decoupling. Together, these results demonstrate the benefit of input token decoupling.

\paragraph{Task-Specific Sensitivity to Visual Token Reduction.}

We further examine whether visual token reduction affects different task types differently by systematically evaluating MSD on DocVQA~\cite{mathew2021docvqa} (OCR-intensive tasks) and VQAv2~\cite{goyal2017vqav2} (natural scene understanding).
As shown in Table~\ref{tab:k_sensitivity}, VQAv2 remains highly stable across different $K$ values, with $\tau$ varying only slightly (e.g., 4.64–4.70). In contrast, DocVQA shows moderate sensitivity to aggressive compression, where increasing $K$ from 32 to 2048 improves $\tau$ from 4.40 to 4.52, suggesting that OCR-intensive tasks benefit from preserving finer-grained visual details.
Nevertheless, smaller $K$ still yields higher overall speedup due to reduced draft latency. Overall, $K=32$ provides a robust trade-off between efficiency and acceptance stability across diverse task types.

\begin{table}[t]
\centering
\scriptsize
\setlength{\tabcolsep}{7pt}
\renewcommand{\arraystretch}{1.0}
\caption{Task-specific sensitivity to visual compression under different retained tokens $K$. Each entry reports $\tau$ / speedup ratio.}
\label{tab:k_sensitivity}

\begin{tabular}{lccc}
\toprule
Task & $K=32$ & $K=512$ & $K=2048$ \\
\midrule
DocVQA  & 4.40 / 2.47$\times$ 
        & 4.45 / 2.45$\times$ 
        & 4.52 / 2.40$\times$ \\
VQAv2   & 4.67 / 2.22$\times$ 
        & 4.64 / 2.16$\times$ 
        & 4.70 / 2.09$\times$ \\
\bottomrule
\end{tabular}

\end{table}

\subsection{Extension to Multi-Image Inputs}

We further examine whether MSD generalizes beyond single-image settings.
Since speculative decoding operates only on the draft model,
the target model remains unchanged when extending to multi-image inputs.

We compare two clustering strategies on MuirBench~\cite{wang2024muirbench}:
(i) \emph{Global Clustering}, which compresses all visual tokens jointly across images,
and (ii) \emph{Per-Image Clustering}, which compresses each image independently before concatenation.

As shown in Table~\ref{tab:multi_image},
Per-Image Clustering achieves a higher acceptance length
($\tau=4.08$) than Global Clustering ($\tau=3.63$),
leading to a higher speedup ratio (2.01$\times$ vs.\ 1.73$\times$).
This indicates that preserving image-specific structures during token reduction
improves draft quality and stabilizes speculative verification.

\begin{table}[t]
\centering
\scriptsize
\setlength{\tabcolsep}{6pt}
\renewcommand{\arraystretch}{1.0}
\caption{Comparison of multi-image token reduction strategies evaluated on MuirBench.}
\label{tab:multi_image}

\begin{tabular}{lcc}
\toprule
Strategy & $\tau$ & SR \\
\midrule
Global Clustering  & 3.63 & 1.73$\times$ \\
Per-Image Clustering  & \textbf{4.08} & \textbf{2.01$\times$} \\
\bottomrule
\end{tabular}

\end{table}

\section{Conclusion}
\label{sec:conclu}

In this paper, we present Multimodal Speculative Decoding (MSD), a novel and effective method for accelerating MLLM inference without compromising accuracy.
 By analyzing the fundamental differences between LLMs and MLLMs, we identify three design principles for multimodal speculative decoding:
 (1) Text and visual tokens differ fundamentally and should be processed separately during drafting.
 (2) Visual inputs to the draft model are highly redundant and can be compressed.
 (3) The draft model must maintain both language modeling ability and visual perception. 
Based on these principles, MSD is designed as a unified framework that decouples input tokens by modality in the draft model, introduces a lightweight visual token-reduction module to compress visual inputs, and adopts a two-stage training strategy that equips the draft model with both capabilities. Extensive experiments across a broad range of challenging multimodal tasks demonstrate that MSD delivers significant inference speedups without sacrificing accuracy.


\appendices

\section{Training Details}
\label{app:train_detail}

We provide detailed training configurations for the draft model used in our MSD framework.

The training objective combines next-token prediction loss $\mathcal{L}_n$ and feature alignment loss $\mathcal{L}_f$ to guide the draft model to generate tokens while aligning intermediate representations with the target model. The total loss is defined as:
\begin{equation}
\mathcal{L} = 0.1 \cdot \mathcal{L}_n + \mathcal{L}_f.
\end{equation}

The next-token prediction loss is formulated as:
\begin{equation}
\mathcal{L}_n = \text{CrossEntropy}(p^i_d, p^i_t),
\end{equation}
where $p^i_d$ and $p^i_t$ denote the predicted token probabilities for the $i$-th token from the draft and target models, respectively.

The feature alignment loss encourages the hidden representations of the draft model to match those of the target model:
\begin{equation}
\mathcal{L}_f = \text{SmoothL1}(f^i_d, f^i_t),
\end{equation}
where $f^i_d$ and $f^i_t$ are the hidden states of the $i$-th token from the draft and target models.

We apply this training objective to widely used MLLMs such as LLaVA-1.5-7B~\cite{liu2023llava1.5}, LLaVA-1.6-7B and LLaVA-1.6-13B~\cite{liu2024llavanext}, and Qwen2.5-VL-7B~\cite{bai2025qwen25vl}.

The draft model is implemented as a lightweight single-layer Transformer. For each target MLLM, its hidden dimension, FFN dimension, and number of attention heads are kept consistent with the corresponding target language model. Detailed configurations are reported in Table~\ref{tab:draft_architecture}.

\begin{table}[t]
    \centering
    
    \caption{Hyperparameters of the single-layer draft Transformer.}
    \label{tab:draft_architecture}
    \resizebox{0.90\columnwidth}{!}{%
    \begin{tabular}{@{}lccc@{}}
        \toprule
        \textbf{Target Model} & \textbf{Hidden Dim.} & \textbf{FFN Dim.} & \textbf{Attention Heads} \\
        \midrule
        LLaVA-1.5/1.6-7B & 4096 & 11008 & 32 \\
        LLaVA-1.6-13B & 5120 & 13824 & 40 \\
        Qwen2.5-VL-7B & 3584 & 18944 & 28 \\
        \bottomrule
    \end{tabular}%
    }
\end{table}

To align with the dataset configuration in ViSpec, we keep the dataset sizes identical across modalities. The text-only instruction-tuning dataset consists of 68,000 randomly sampled dialogues from ShareGPT. For the visual instruction-tuning dataset, we also use 68,000 samples, where each sample is extended by prompting the target VLM to generate long, detailed responses based on the LLaVA-Instruct-150K data.

The draft models are trained for 20 epochs using the progressive multimodal data mixing strategy (defined in Eq.~\ref{eq:progressive_data}) on four NVIDIA H200 GPUs, and the complete training process takes approximately 6.1 hours in total (24.4 H200 GPU-hours). We use a batch size of 4 and an initial learning rate of $5 \times 10^{-5}$, scheduled using a cosine annealing strategy. Unless otherwise specified, all evaluations are conducted on NVIDIA H200 GPUs.

For speculative decoding, we adopt a draft tree with a depth of 6 and a total of 60 draft tokens. This configuration balances exploration capacity and computational efficiency, and is used consistently across all models and experiments.

\begin{table}[t]
    \centering
    
    \caption{Sensitivity of average acceptance length to K-Means initialization seeds on VQAv2 with LLaVA-1.6-7B.}
    \label{tab:kmeans_seed_sensitivity}
    \resizebox{0.92\columnwidth}{!}{%
    \begin{tabular}{@{}lccccc@{}}
        \toprule
        \textbf{Random Seed} & 0 & 1 & 2 & 42 & 1234 \\
        \midrule
        \textbf{Average Acceptance Length} & 4.66 & 4.67 & 4.66 & 4.68 & 4.67 \\
        \bottomrule
    \end{tabular}%
    }
\end{table}

Across five random initializations in Table~\ref{tab:kmeans_seed_sensitivity}, the average acceptance length ranges from 4.66 to 4.68, showing low sensitivity to K-Means initialization.

\section{Baseline Methods}
\label{app:baselines}

In this section, we describe the baseline methods used in our experiments.

(1) \textbf{Baseline}: The feature-level SD method described in Sec.\,\ref{sec:baseline}. It retains the original shifted feature construction and serves as the direct reference for evaluating the gains introduced by MSD. (2) \textbf{Lookahead}: A training-free SD approach for LLMs~\cite{fu2024lade}, which generates n-gram sequences using Jacobi iteration during drafting and verifies them with the target model. (3) \textbf{Medusa}: A parallel SD method that attaches multiple prediction heads to the target model, enabling simultaneous generation of multiple future tokens~\cite{cai2024medusa}. Both Lookahead and Medusa are extended to the multimodal setting by adapting them to LLaVA for comparison. Both Medusa and Lookahead use the default hyperparameter settings provided in their official repositories. (4) \textbf{Text\_Only}: Following Gagrani et al.~\cite{gagrani2024spd_mllm}, this draft model uses only the text part of the input question and ignores visual features. (5) \textbf{ViSpec}: A vision-aware SD method that uses a lightweight vision adaptor to compress visual tokens and a global visual feature to uniformly augment text tokens.

For a fair comparison, Lookahead follows its official training-free configuration. Medusa, ViSpec, and the feature-level Baseline use the same visual instruction data, while Text\_Only uses an equally sized text set. For all trainable methods, we use the same number of training samples, training epochs, and tuning budget. During evaluation, all methods share the same target model, test samples, hardware, batch size, temperature, output length, and timing protocol, with their actual drafting and verification costs included in the reported speedups. All tree-based methods further use a draft-tree depth of 6 and 60 candidate tokens, giving them the same search budget and structural upper bound per verification step.

\section{Speedup Ratio Analysis}
\label{app:speedup}

The speedup ratio (SR) measures the relative acceleration of speculative decoding compared to vanilla autoregressive decoding.

Given an average acceptance rate \( \alpha \in [0,1] \) and a speculative length \( \gamma \), the expected number of accepted tokens per verification step is defined as~\cite{chen2024magicdec}:
\begin{equation}
    \omega(\gamma, \alpha) := \mathbb{E}(\# \text{generated tokens}) = \frac{1 - \alpha^{\gamma+1}}{1 - \alpha}.
\end{equation}

Including prefilling time \( T_{prefilling} \) and output length \( N \), the speedup ratio is calculated as:

\begin{equation}
  \begin{aligned}
    \text{SR} &=\frac{\text{Time(Normal Decoding)}}{\text{Time(Speculative Decoding)}} \\
  &= \frac{N \cdot T_p + T_{prefilling}}{ N \cdot \left( \frac{\gamma \cdot T_q + T_v(\gamma)}{\omega(\gamma, \alpha)} \right) + T_{prefilling}},
  \end{aligned}
  \label{eq:speedup}
\end{equation}

where \( T_p \) is the time the target model takes to generate one token, and \( T_q \) is the time for the draft model.
\( T_v(\gamma) \) denotes the verification time of the target model for \( \gamma \) tokens.

\textit{Effect of Input and Output Lengths.} We evaluate MSD across tasks with different input and output lengths using LLaVA-1.6-7B at $T=0$. To isolate the effect of sequence length, we select AI2D, SQA\_Image, and TextVQA, whose average acceptance lengths are all approximately 4.4.

\begin{table}[t]
    \centering
    
    \caption{Effect of input and output lengths on MSD efficiency. All values are averages over evaluation samples. All throughput and speedup measurements are conducted on an NVIDIA A100 GPU.}
    \label{tab:length_speedup}
    \resizebox{\columnwidth}{!}{%
    \begin{tabular}{@{}lccccc@{}}
        \toprule
        \textbf{Task} & \textbf{Input Tokens} & \textbf{Output Tokens} & $\boldsymbol{\tau}$ & \textbf{Throughput (tok/s)} & \textbf{Speedup} \\
        \midrule
        AI2D & 1923.7 & 213.2 & 4.41 & 88.23 & 2.82$\times$ \\
        SQA\_Image & 1865.8 & 141.9 & 4.41 & 84.44 & 2.77$\times$ \\
        TextVQA & 2336.9 & 81.6 & 4.39 & 66.43 & 2.35$\times$ \\
        \bottomrule
    \end{tabular}%
    }
\end{table}

As shown in Table~\ref{tab:length_speedup}, the three tasks have nearly identical acceptance lengths, so their efficiency differences mainly reflect their input and output lengths. Consistent with Eq.~\ref{eq:speedup}, shorter inputs reduce the unaccelerated prefilling cost, while longer outputs increase the fraction of computation accelerated by speculative decoding. AI2D therefore reaches 88.23 tok/s and a 2.82$\times$ speedup, while TextVQA, with the longest input and shortest output, reaches 66.43 tok/s and 2.35$\times$.

\section{Analysis of Speedup Ratio Differences Between MLLMs and LLMs}
\label{appendix:speedup_gap}

This appendix provides a detailed explanation of why MSD achieves comparable average acceptance lengths between MLLMs and LLMs but still exhibits a lower overall speedup ratio in multimodal benchmarks.

\paragraph{Different Input–Output Length Characteristics}
The key factor lies in the substantial disparity between input and output lengths in MLLM versus LLM evaluation settings.  
In typical LLM benchmarks (e.g., MT-Bench~\cite{zheng2023mtbench}, HumanEval~\cite{chen2021humaneval}, GSM8K~\cite{cobbe2021gsm8k}), the average input and output lengths are approximately 428 and 388 tokens, respectively.  
In contrast, MLLM benchmarks (e.g., ChartQA~\cite{masry2022chartqa}, TextVQA~\cite{Singh2019textvqa}, VQAv2~\cite{goyal2017vqav2}, HallusionBench~\cite{guan2024hallusionbench}) involve much longer textual–visual inputs (around 2024 tokens) but much shorter outputs (around 85 tokens).  
This imbalance fundamentally limits the amount of acceleration that speculative decoding can exploit.

\paragraph{Implications for Speedup Ratio}
Speculative decoding primarily accelerates generation during the output phase, so shorter outputs in MLLM tasks reduce the effective region where speedup can be realized.  
Consequently, even though MSD yields similar or larger acceptance lengths than LLM speculative decoding, the overall multiplicative speedup ratio is inherently bounded in MLLM benchmarks.

\paragraph{Summary}
These characteristics explain why MSD can match LLMs in acceptance length while still showing lower end-to-end speedup ratios on multimodal tasks.

\begin{figure}[!t]
  \centering
  \includegraphics[width=0.47\textwidth]{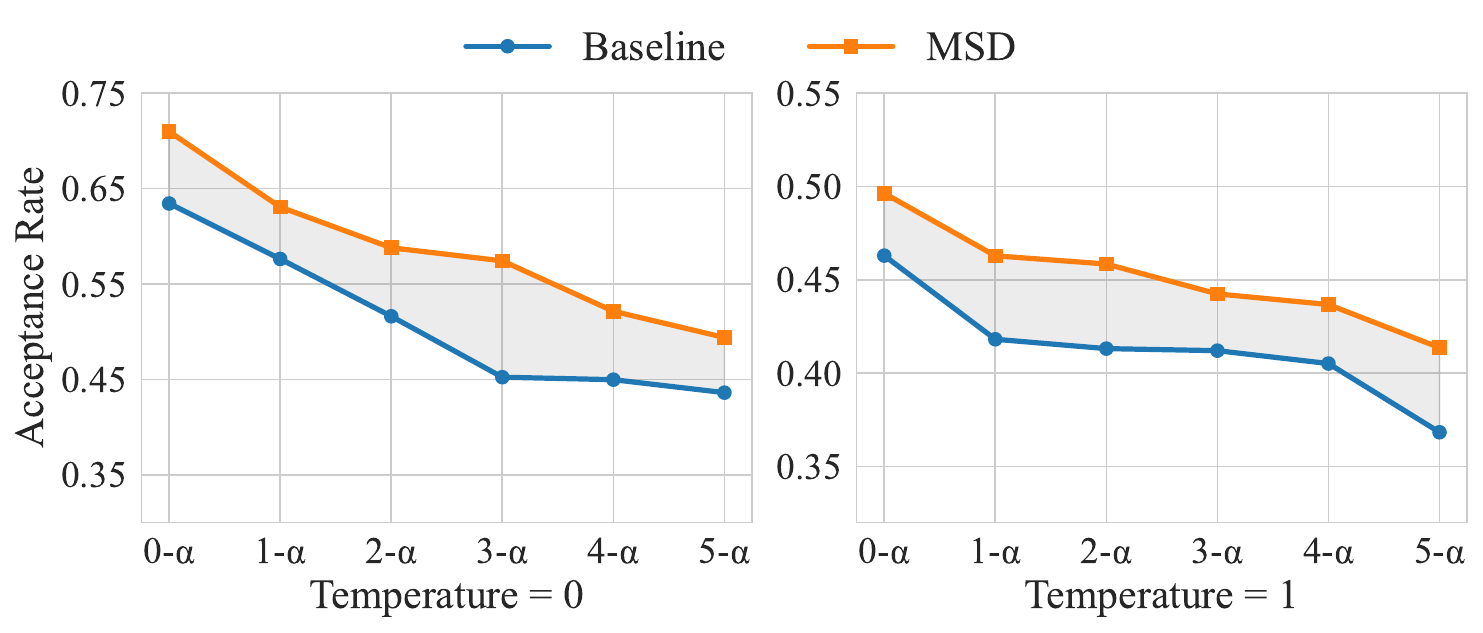}
  \caption{
Comparison of acceptance rates between Baseline and MSD across different temperatures on TextVQA with LLaVA-1.6-7B.
  }
  \label{fig:n_alpha}
\end{figure}
\section{Analysis of Acceptance Rate}
\label{app:accerate}
The acceptance rate $\alpha$ represents the ratio of accepted to candidate tokens during drafting.
We use chain drafts, where the draft model generates one token per forward pass. Since the draft model predicts features autoregressively, errors may accumulate across steps.
Following previous work~\cite{li2024eagle}, we measure the acceptance rate as $n\text{-}\alpha$, where $n\text{-}\alpha$ is the acceptance rate of the $n$-th predicted token during drafting.
As shown in Figure~\ref{fig:n_alpha}, MSD achieves a higher acceptance rate than Baseline across different $n$ and temperature settings.

\section{Latency Overhead of Visual Token Reduction}

\begin{table}[t]
\centering
\caption{Latency overhead of the visual token reduction module at different visual token resolutions, averaged over the ChartQA benchmark on an NVIDIA H200 GPU.}
\label{tab:clustering_latency}
\resizebox{\columnwidth}{!}{
\begin{tabular}{cccc}
\toprule
\textbf{\# Visual Tokens} & \textbf{Clustering (ms)} & \textbf{Total Drafting (ms/sample)} & \textbf{Overhead (\%)} \\
\midrule
576   & 0.93 & 620.4  & 0.15\% \\
1917  & 1.28 & 1163.6 & 0.11\% \\
\bottomrule
\end{tabular}
}
\end{table}


\label{app:clustering_latency}
Table~\ref{tab:clustering_latency} reports the runtime overhead introduced by the visual token reduction module. 
The clustering step exhibits linear complexity $O(n)$ with respect to the number of visual tokens and remains highly efficient in practice. 
For 576 visual tokens, the clustering operation takes 0.93\,ms, accounting for only 0.15\% of the total drafting time per sample.
Even at a higher resolution with 1917 visual tokens, the overhead increases marginally to 1.28\,ms (0.11\%). 
The slight decrease in relative overhead at higher resolutions is expected, since the transformer-based draft computation grows super-linearly with sequence length, while the clustering step scales linearly.
These results demonstrate that the clustering module introduces negligible latency relative to the transformer-based draft computation, and thus does not constitute a computational bottleneck in MSD.

\begin{figure}[t]
    \centering
    \includegraphics[width=0.40\textwidth]{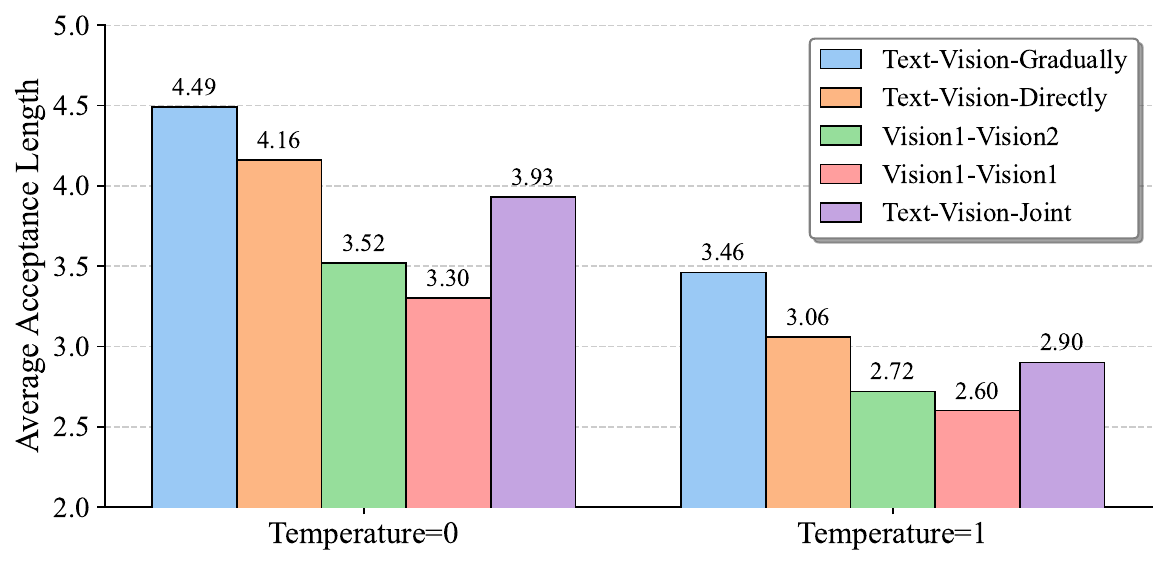}
    \caption{
Impact of draft model training strategies evaluated on ChartQA with LLaVA-1.6-7B.
    }
    \label{fig:datasets}
\end{figure}

\section{Ablation on Two-Stage Training Strategy}
\label{ablation: training strategy}

MSD adopts a two-stage training strategy in which the draft model is first trained on a text-only dataset and then gradually transitioned to a visual dataset. We compare our Text-Vision-Gradually approach with three alternatives: Vision1-Vision1, Vision1-Vision2, and Text-Vision-Directly. As shown in Figure~\ref{fig:datasets}, text-vision strategies consistently surpass vision-only ones, indicating that the gains come not only from data diversity but also from the strong language modeling learned in Stage~1.
We also include a \textbf{Text-Vision-Joint} variant, which mixes text and visual data at a fixed 1:1 ratio in a single-stage setup. To ensure a fair comparison, this variant uses the same total number of training samples as the other strategies. While it performs better than vision-only training, it still falls short of the transition-based strategies, suggesting that directly combining modalities is less effective than introducing them through a staged process.
Text-Vision-Directly preserves the same two training stages but switches abruptly from text to visual data, thereby isolating the effect of the progressive transition. Under the same data budget and training settings, Text-Vision-Gradually performs best at both temperatures, showing that its gain comes from the progressive schedule itself rather than simply using both text and visual data.

\section{Verification of Lossless Decoding}
\label{app:lossless_verification}

\begingroup
\setlength{\intextsep}{3pt}
\begin{table}[!b]
    \centering
    
    \caption{Task scores of LLaVA-1.6-7B with vanilla decoding and MSD under greedy decoding ($T=0$).}
    \label{tab:lossless_verification}
    \resizebox{0.90\columnwidth}{!}{%
    \begin{tabular}{@{}lcccc@{}}
        \toprule
        \textbf{Method} & \textbf{ChartQA} & \textbf{AI2D} & \textbf{Hallusion} & \textbf{VQAv2} \\
        \midrule
        LLaVA-1.6-7B & 0.36 & 0.68 & 0.40 & 0.81 \\
        LLaVA-1.6-7B + MSD & 0.36 & 0.68 & 0.40 & 0.81 \\
        \bottomrule
    \end{tabular}%
    }
\end{table}
\endgroup

All MSD-specific operations affect only the proposal distribution $q$. The target model retains the complete visual input and evaluates each candidate under $p$ from its original prefix, while tree masks and position IDs only batch candidate branches without changing their conditioning context. Following standard speculative decoding~\cite{leviathan2023spd}, accepting a proposal $y\sim q$ with probability $\min(1,p(y)/q(y))$ and sampling after rejection from the normalized residual distribution $[p-q]_+$ preserves $p$ at every step. Under greedy decoding, any mismatch is replaced by the target prediction. Importantly, neither compressed visual features nor draft hidden predictions enter the target-model probability computation. Therefore, the MLLM-specific components change only how candidates are proposed, while the verification path remains identical to standard speculative decoding. Thus, MSD changes proposal efficiency without changing the final output distribution.

Table~\ref{tab:lossless_verification} reports identical scores on all four benchmarks, further validating the lossless implementation.



\bibliographystyle{IEEEtran}
\bibliography{main_revision2}

\vfill

\end{document}